\documentclass[10pt,conference,compsocconf]{IEEEtran}

\usepackage{url}
\usepackage{textcomp}
\usepackage{graphicx}	

\begin{document}
\title{Segmentation of Roads in Satellite Images \\ using specially modified U-Net CNNs}

\author{
 Jonas Bokstaller, Yihang She, Zhehan Fu, Tommaso Macrì\\
 bojonas@ethz.ch, yihshe@ethz.ch, zhehfu@ethz.ch, macrit@ethz.ch
}

\maketitle

\begin{abstract}
The image classification problem has been deeply investigated by the research community, with computer vision algorithms and with the help of Neural Networks.
The aim of this paper is to build an image classifier for satellite images of urban scenes that identifies the portions of the images in which a road is located, separating these portions from the rest. Unlike conventional computer vision algorithms, convolutional neural networks (CNNs) provide accurate and reliable results on this task. Our novel approach uses a sliding window to extract patches out of the whole image, data augmentation for generating more training/testing data and lastly a series of specially modified U-Net CNNs. This proposed technique outperforms all other baselines tested in terms of mean F-score metric.
\end{abstract}

\section{Introduction}
The task of image segmentation consists of choosing the right label for each pixel such that the image is segmented into different parts like in this work a satellite image into road segments and background (e.g buildings, trees, grass). This task is more important then ever because of the recent developments of autonomous driving vehicles, which requires precise guidance for save driving.

Advances in computational infrastructure and breakthroughs in artificial neural network architectures (ANNs) are emerging. These lead to the use of machine learning to tackle this partitioning task. In order to process the amount of information in a single image efficiently, we need to use convolutional neural networks (CNNs) \cite{cnn} because they split, downscale and upscale the image while applying different filters in an efficient way. This enables the use of such networks for real time image segmentation with high accuracy.

As training data we have satellite images as well as the corresponding ground-truth map which labels each pixel as road or background. The task of the paper is to predict the label of new satellite images in chunks of 16x16 pixels. The test images have dimensions 608x608 pixels, therefore we have to predict 1444 chunks per image. In order to compare the performance of different models and approaches, mean $F_1$ score is used. The road label is assigned to a chunk if the proportion of road pixels exceed the threshold of 25\% in the chunk.
\begin{figure}[h]
\includegraphics[width=\linewidth]{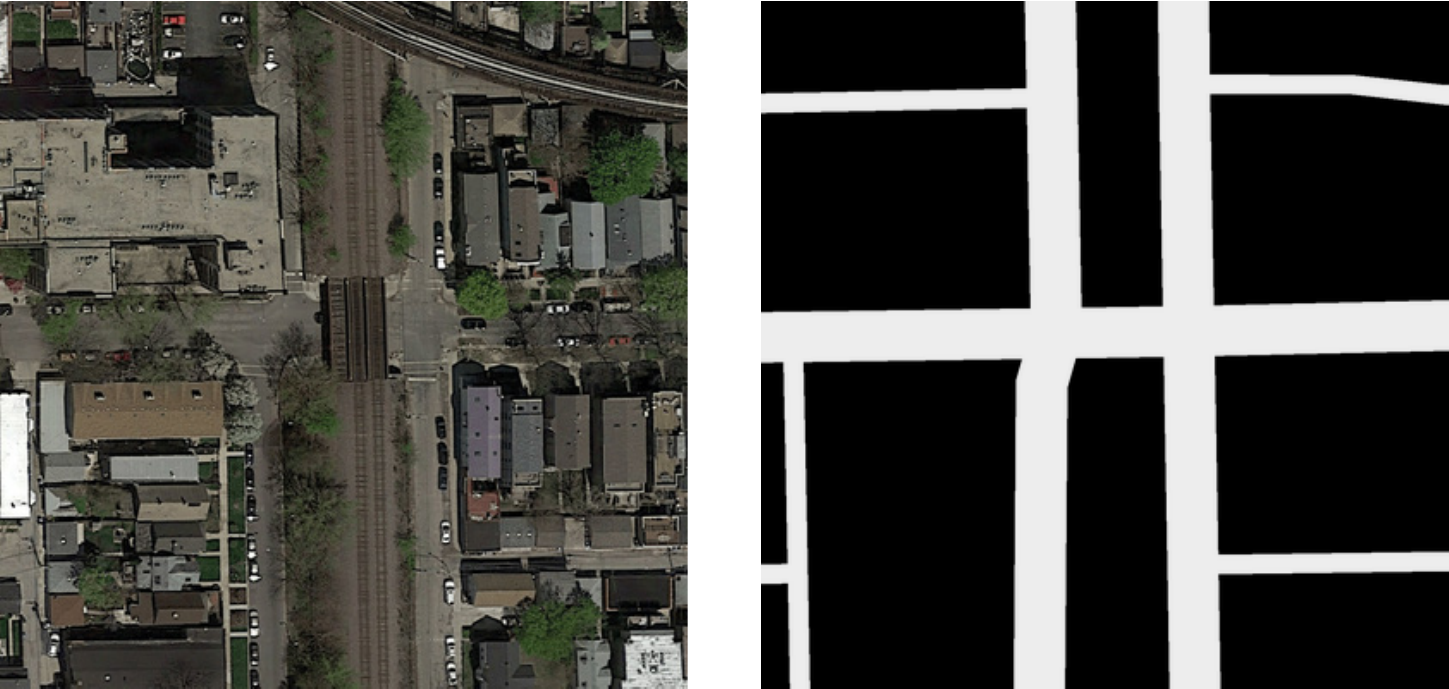}
\caption{Example from the provided training data. On the left there is the input satellite image and on the right its respective ground-truth.}
\end{figure}

This paper gives the reader a brief overview over the different methods we implemented, the difficulties we had encountered during testing and our final solution which consists of a set of specially modified U-Net CNNs.

\section{Models and Methods}

\subsection{General Workflow}
Given the training set of 100 images and corresponding masks with size $400\times400$, the task requires classifying the $16\times16$ patches of the test set into either road or background. To conduct such classification, we have tested three baseline models:
\begin{enumerate}
    \item The sliding window approach, which conducts a binary classification for each patch with the given window size using convolutional neural networks.
    \item The segmentation approach based on U-Net to generate pixel-level classification. 
    \item Transfer learning strategy for segmentation. The encoder of U-Net is adapted from the base model of MobileNetV2. Pre-trained parameters on ImageNet were taken for initialization.
\end{enumerate}
Besides, several data preprocessing techniques and training strategies were applied. Our final solution involves averaging outputs of three models based on U-Net, which result from improvements on network architecture of original U-net model. This combination has significantly improved F-score on both validation and test set. 

\subsection{Data Pre-processing and Augmentation}
As the number of aerial images in training set is relatively small (with only 100 images and corresponding masks), we generated 900 patches of $256\times256$ in size from them. These patches are used throughout the second and third baseline solution and the final one. Furthermore, image data augmentation techniques were used to overcome the impacts of limited training data size. We augmented the images by rotation ($rotation\_range=0.2$), shifting the width and height of the image center ($width\_shift\_range=0.2$, $height\_shift\_range=0.2$), shearing and zooming the image ($shear\_range=0.2$, $zoom\_range=0.2$), and random flipping in both horizontal and vertical directions. The corresponding masks were also transformed with the same operations. 

\subsection{Network Architecture}
\subsubsection{Baseline solution 1: sliding-window approach}
This approach is based on the work of Gouk et al.\cite{slidingWindow} The main idea behind this approach is to classify each patch with padding region around it to provide the model with enough context information. Then we use this window and scan over the image in order to classify each patch. The patch plus the padding is then fed into a convolutional neural network with 5 layers in total. To avoid overfitting, Leaky ReLu was used as activation function, and Dropout was added with a probability of 0.5.\cite{srivastava2014dropout}
\begin{figure}[h]
\includegraphics[width=\linewidth]{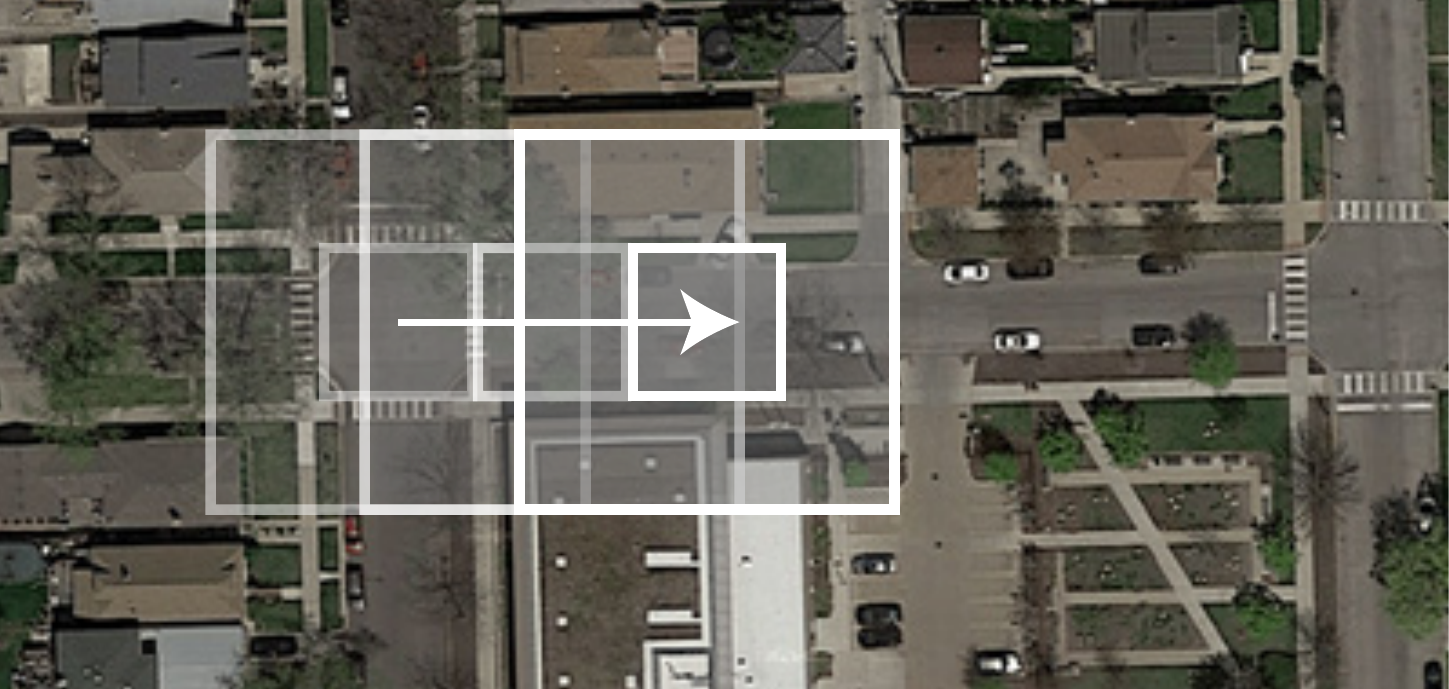}
\caption{Patch with padding around it called sliding window approach. The padding is used to provide the model with enough contextual information. The stride of the sliding window is its own size, namely 16x16.}
\end{figure}

To further avoid overfitting and  to increase the number of windows as training data, we used a data augmentation strategy different from the aforementioned one. First, a patch is randomly selected as training instance and the window inside that patch is chosen. Then it underwent random flipping and rotation so that  the model's generalization ability is further enhanced. If the patch is at the border, the padding is just a mirrored image around its edge.

\subsubsection{Baseline solution 2: U-Net}

\begin{figure*}[ht]
\vskip 0.2in
\begin{center}
\centerline{\includegraphics[width=140mm]{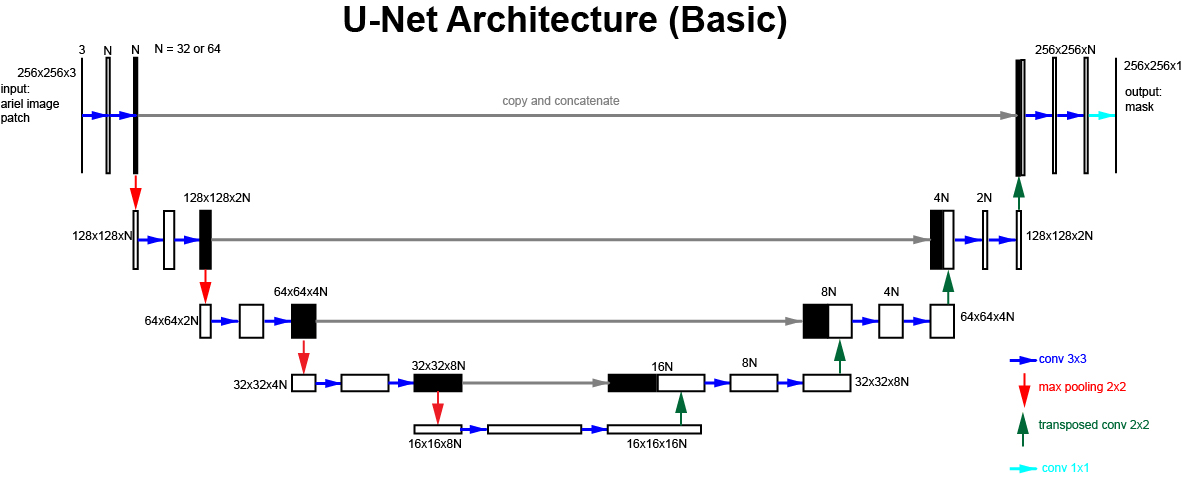}}
\caption{The structure of the modified U-net architecture.
}
\label{fig:unetbasic}
\end{center}
\vskip -0.2in
\end{figure*}

We implemented the original U-Net architecture\cite{ronneberger2015u} on this task. (see Figure \ref{fig:unetbasic}) This is a small and easily trainable fully convolutional neural network designed originally for binary segmentation of biomedical images. The network is consisted of two paths, the encoding and decoding paths. In the first half or the encoding path, operations are 2D convolutions followed by a non-linear activation function, ELU.\cite{clevert2015fast} Feature map size is retained in each convolution by padding. Then max pooling is used to reduce the feature map in the encoding path. After these operations, number of feature channels is doubled.
Then the U-Net uses an expansion path (decoding path) to up-sample the feature maps and eventually create a segmentation mask same in size as input image. This path consists of sequences of up sampling in the form of transpose convolutions, followed by concatenations with feature maps copied from the encoding path. Finally, a convolution of kernel size (1,1) and a Softmax function is applied. This maps each entry in input aerial image to one of two classes, road or non-road.

\subsubsection{Baseline solution 3: U-Net with MobileNetV2 trained on ImageNet}
In order to extract robust features, We further applied transfer learning, specifically using pre-trained parts of a lightweight model, MobileNetV2\cite{sandler2018mobilenetv2} as encoder path of U-Net. This is especially applicable in this task because it has a rather limited training set, and the method does not require much computational resource. The base model to transfer, MobileNetV2 is trained on a large dataset of another task, in this case the ImageNet datset\cite{deng2009imagenet}, so that the weights of its convolutional layers (especially the first few ones) are more robust in extracting features that can be applied to the road extraction task. The fundamental architecture of the baseline model is based on U-Net. We adapted its encoder from the convolutional layers of pre-trained MobileNetV2. The output layer of the encoder is the layer $block\_13\_expand\_relu$ of MobileNetV2. Four output layers of MobileNetV2 ($input\_image$, $block1\_expand\_relu$, $block\_3\_expand\_relu$, $block\_6\_expand\_relu$) were concatenated to the transposed convolutional layers of the encoder to build skip connections. 

\subsubsection{Final solution: improvements on U-Net}
\begin{figure}[ht]
\vskip 0.2in
\begin{center}
\centerline{\includegraphics[width=\columnwidth]{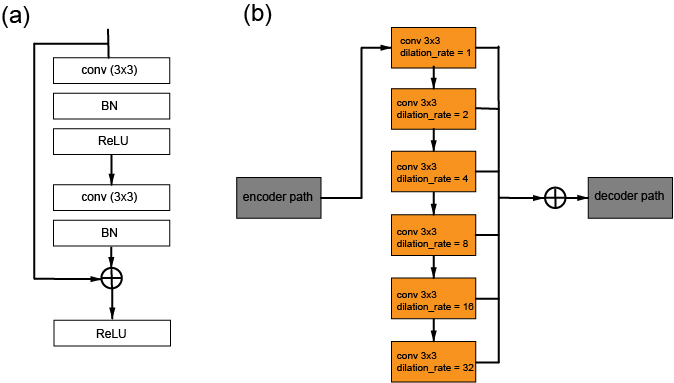}}
\caption{The major improvements on U-Net architecture: (a) addition of identity mapping before final nonlinear activation in each convolution block, as in ResNet; (b) Substituting the original bottleneck with dilated convolutions.}
\label{fig:improvements}
\end{center}
\vskip -0.2in
\end{figure}

We made two major improvements on U-Net architecture and implemented three different models (Figure \ref{fig:improvements}). The final submitted result was generated through averaging the values of masks generated by three models (after Softmax function) and classification was done with a threshold of 0.25.

First, we added local skip connections (similar to that of ResNet\cite{he2016identity}) to the convolution blocks in both the encoding and decoding pathways of U-Net. This is done through addition of the current layer input's identity mapping to its output values. Each modified convolution block takes the following general form: 
$$\begin{array}{l}
\mathbf{y}_{l}=h\left(\mathbf{x}_{l}\right)+\mathcal{F}\left(\mathbf{x}_{l}, \mathcal{W}_{l}\right) \\
\mathbf{x}_{l+1}=f\left(\mathbf{y}_{l}\right)
\end{array}$$
where $\mathbf{x}_{l}$ and $\mathbf{x}_{l+1}$ are $l$-level's input and output tensors, $\mathcal{F}(\cdot)$ is the residual function, which is the original convolutions followed by batch normalization and nonlinear activation. $f\left(\mathbf{y}_{l}\right)$ is the activation function, and $h\left(\mathbf{x}_{l}\right)$ is the identity mapping function, which is a simple 2D convolution with kernel size 1 in our solution. 
This added operation addresses 	the gradient degradation problem by letting it backpropagate through this additional identity mapping function. It also facilitates training, and utilizes features of lower semantic information extracted in the previous layers.

The second improvement is substitution of the central bottleneck in U-Net with a series of dilated convolution operations. The summation of these convolutions are fed into the decoder path. Previous work \cite{piao2019accuracy}, \cite{yu2015multi} reports that adding dilated convolution modules can solve the problem of degraded picture resolution. This is achieved through expanded receptive fields, thus maintaining per-pixel level classification accuracy while generating large-scale feature maps with rich context information.

In all, we implemented three models based on the above improvements. They are summarized in the table below.
\begin{table}[ht]
\caption{Models used in final solution} 
\centering 
\begin{tabular}{c c c c} 
\hline\hline 
Model name & Description \\ [0.5ex] 
\hline 
unet-32 & 32 channels on the first layer, local skip connections \\ 
unet-64 & 64 channels on the first layer, local skip connections \\
unet-dilated & parallel dilated convolutions as bottleneck, local skip connections \\ [1ex] 
\hline 
\end{tabular}
\label{table:models} 
\end{table}

\subsection{Loss function and evaluation metrics}
The models were trained to optimize the smoothed dice loss, defined as follows:
$$L=1-\frac{2|y \cap \hat{y}|+\epsilon}{|y|+|\hat{y}|+\epsilon} $$
where $y$ is the mask values predicted by model, $\hat{y}$ is the ground-truth, and $\epsilon$ is the smoothing coefficient, selected as 1. Model performance is evaluated based on the following metrics: IoU (intersection over union) and F-1 score ($t_p$ is true positive, $f_p$ and $f_n$ are false positive and false negative ratios):
$$I o U = \frac{|y \cap \hat{y}|}{|y \cup \hat{y}|} $$
$$F_1 = \frac{\mathrm{t_p} }{\mathrm{t_p}+\frac{1}{2}(\mathrm{f_p}+\mathrm{f_n})} $$

\subsection{Training}

We split the data into training ($80\%$) and validation set ($20\%$). The initial learning rate is 0.0001 and it will be reduced with a factor of 0.5 if the validation loss is less than 0.0002 after every 5 epochs. The model was built in Keras\cite{chollet2015keras} and trained with a batch size of 8 with 100 epochs in maximum. When training the unet-32 and unet-64 models (described in Table \ref{table:models}), dropout was added after each concatenation with probability 0.2. Batch normalization was performed to accelerate training (as illustrated in Figure \ref{fig:improvements}), in the position after each 2D convolution and before nonlinear activation functions. Baseline solutions were trained using Google Colaboratory\cite{bisong2019google} and final solution models were trained on a cluster using one NVIDIA\textregistered V100 GPU and 28 Intel\textregistered Xeon\textregistered E5-2690 v4 CPUs.

\section{Results}
\subsection{Key results from baselines}
\subsubsection{Sliding window approach results}
Hyperparameters were fine-tuned for this approach and mean F-score on test set was around 0.85.
\subsubsection{Transfer learning results}
The IoU on the validation set finally achieves 0.7991, which cannot meet typical standards of road extraction. Given that the outcome of transfer learning also depends on the similarity between the pretrained dataset and the specific task, the trained model conditioned on ImageNet might only be able to find a suboptimal parameters for road segmentation.
\subsection{Final solution results}
We observed a certain degree of improvement on all three models proposed above than the traditional U-Net architecture. Detailed performance metrics on validation and test set are listed below:

\begin{table}[ht]
\caption{Performance metrics of proposed models} 
\centering 
\begin{tabular}{c c c c} 
\hline\hline 
Model & val\_loss & val\_f1 & test\_f1 \\ [0.5ex] 
\hline 
unet-32 & 0.0826 & 0.9174 & 0.8780 \\ 
unet-64 & 0.0368 & 0.9632 & 0.8959 \\
unet-dilated & 0.0334 & 0.9666 & 0.9015 \\ [1ex] 
\hline 
\end{tabular}
\label{table:perf} 
\end{table}

Averaging the values of masks predicted by these models yield higher $F_1$ score on test set (0.9027), which was chosen for submission.

\section{Discussion}
In this work we first discovered that by using encoder-decoder architecture such as U-net, semantic information in different levels are effectively coupled to give clearer and more accurate results than a classification of each patch with a simple 5-layer CNN. By further substituting bottleneck of U-Net with parallel dilated convolution structure and adding residual-like skip connections, we largely exceeded performance of baseline solutions. However, there is still room of improvement for each of the models proposed. The gap between score on test set and validation set suggests that the problem of lacking training data is still not resolved. Possible candidate methods are image sharpening with  filters  and color jittering. Also, different selections of layer parameters transferred  into  U-Net  as  well  as  different  tasks  the  base model  is  trained  on  needs  to  be  explored  further.  This  can affect  overall  model  performance  and  the  amount  of  time required to train until convergence.

\bibliographystyle{IEEEtran}
\bibliography{segmentation-paper}
\end{document}